%% file: root.tex
\PassOptionsToPackage{cleveref}{jmlrutils}

\documentclass{l4dc2027}
\jmlrproceedings{}{Preprint. Under review.}\jmlrvolume{}\jmlrpages{}\jmlryear{}

\title[Learning Dynamics of DLOs through Coupled Oscillators for Bimanual Manipulation]{Learning the Dynamics of Deformable Linear Objects Through Coupled Oscillators for Bimanual Robotic Manipulation}
\usepackage{times}

\usepackage[nolist,nohyperlinks]{acronym}

\usepackage{textcomp}
\usepackage{cancel}
\usepackage{booktabs} %
\usepackage{tikz}
\usetikzlibrary{positioning,shapes.geometric,arrows.meta}

\usepackage{enumerate}   %
\usepackage{multirow} %
\usepackage{wrapfig}
\usepackage{placeins} %

\newif\ifcompactfloats
\compactfloatstrue
\edef\restorefloatspacing{%
  \noexpand\setlength{\textfloatsep}{\the\textfloatsep}%
  \noexpand\setlength{\floatsep}{\the\floatsep}%
  \noexpand\setlength{\intextsep}{\the\intextsep}%
  \noexpand\setlength{\dbltextfloatsep}{\the\dbltextfloatsep}%
  \noexpand\setlength{\dblfloatsep}{\the\dblfloatsep}%
  \noexpand\setlength{\abovecaptionskip}{\the\abovecaptionskip}%
  \noexpand\setlength{\belowcaptionskip}{\the\belowcaptionskip}%
}
\usepackage{etoolbox}
\makeatletter
\let\originalmakecaption\@makecaption
\appto\restorefloatspacing{\let\@makecaption\originalmakecaption\compactfloatsfalse}
\patchcmd{\@makecaption}{\vskip 10pt}{\vskip\abovecaptionskip}{}%
  {\PackageError{floatspacing}{Could not patch caption spacing}{Check the document class.}}
\patchcmd{\@makecaption}{\settowidth{\labelwidth}{#1:}}%
  {\setlength{\topsep}{0pt}\setlength{\partopsep}{0pt}%
   \setlength{\parsep}{0pt}\settowidth{\labelwidth}{#1:}}{}%
  {\PackageError{floatspacing}{Could not patch multiline captions}{Check the document class.}}
\apptocmd{\@makecaption}{\vskip\belowcaptionskip}{}%
  {\PackageError{floatspacing}{Could not patch caption end spacing}{Check the document class.}}
\makeatother
\AtBeginEnvironment{table}{\ifcompactfloats\setlength{\abovecaptionskip}{0pt}\setlength{\belowcaptionskip}{4pt}\fi}
\AtBeginEnvironment{table*}{\ifcompactfloats\setlength{\abovecaptionskip}{0pt}\setlength{\belowcaptionskip}{4pt}\fi}

\usepackage{siunitx}

\definecolor{mplblue}{HTML}{1F77B4}
\definecolor{mplorange}{HTML}{FF7F0E}
\definecolor{mplgreen}{HTML}{2CA02C}
\definecolor{mplolive}{HTML}{BCBD22}
\definecolor{mplpink}{HTML}{CC79A7}
\definecolor{oiblue}{HTML}{0072B2}
\definecolor{oivermillion}{HTML}{D55E00}
\definecolor{oigreen}{HTML}{009E73}

\definecolor{lightorange}{HTML}{FFB347}
\definecolor{obsspace}{HTML}{567299}
\definecolor{latspace}{HTML}{AC7C00}

\crefname{figure}{Fig.}{Figs.}
\crefname{table}{Table}{Tables}
\crefname{section}{Sec.}{Secs.}
\crefname{equation}{Eq.}{Eqs.}

\newtheorem*{proposition*}{Proposition}
\newtheorem{assumption}{Assumption}

\newcommand{\loss}{\ell}

\def\BibTeX{{\rm B\kern-.05em{\sc i\kern-.025em b}\kern-.08em
    T\kern-.1667em\lower.7ex\hbox{E}\kern-.125emX}}

\ifanonsubmission
\author{\Name{Anonymous Authors}}
\else
\author[López-Broceño$^{1}$ Stölzle$^{1,2}$ Feliu-Talegon$^{1}$ Rus$^{2}$ Della Santina$^{1}$]{%
 \Name{Iván López-Broceño\nametag{$^{1}$}} \Email{ivanlpzbrc01@gmail.com}\\
 \Name{Maximilian Stölzle\nametag{$^{1,2}$}} \Email{mstolzle@mit.edu}\\
 \Name{Daniel Feliu-Talegon\nametag{$^{1}$}} \Email{d.feliutalegon@tudelft.nl}\\
 \Name{Daniela Rus\nametag{$^{2}$}} \Email{rus@csail.mit.edu}\\
 \Name{Cosimo {Della Santina}\nametag{$^{1}$}} \Email{c.dellasantina@tudelft.nl}\\
 \addr $^{1}$Cognitive Robotics, Delft University of Technology, 2628 Delft, CD, Netherlands\\
 \addr $^{2}$Computer Science and Artificial Intelligence Laboratory, Massachusetts Institute of Technology, Cambridge, MA 02139, USA%
}
\fi

\begin{document}

\maketitle
\input{acronyms}
\input{abstract}


\input{sections/1-introduction}
\input{sections/2-problem_statement}
\input{sections/3-model_learning}

\input{sections/4-control}
\input{sections/5-simulation}
\input{sections/6-real_experiments}

\input{sections/7-conclusion}

\clearpage
\restorefloatspacing
\ifanonsubmission\else
\input{sections/acknowledgements}
\fi
\input{sections/9-declaration-of-ai}

\bibliography{references}
\vfill
\pagebreak

\input{sections/8-appendices}

\end{document}

%% file: acronyms.tex
\begin{acronym}
\acro{DLO}{deformable linear object}
\acro{MLP}{multilayer perceptron}
\acro{CON}{coupled oscillator network}
\acro{FBCON}{floating-base coupled oscillator network}
\acro{NODE}{neural ordinary differential equation}
\acro{LNN}{Lagrangian neural network}
\acro{RMSE}{root mean squared error}
\end{acronym}

%% file: abstract.tex
\begin{abstract}

Many learning-based approaches for deformable linear objects (DLOs) rely on simplified deformation models that neglect dynamic effects and the mechanical response of the objects, limiting their applicability to lightweight, compliant objects and slow motions.
Inspired by recent advances in learning physically structured dynamics with coupled oscillator networks (CONs), we propose $\Phi$-CON for learning force-driven DLO dynamics.
The model combines a floating-base coupled oscillator network (FBCON), whose latent potential energy depends only on the DLO's shape, with a learned affine diffeomorphism $\Phi$ between observed positions along the DLO and latent coordinates.
Its shape dynamics are stable by construction, and its Lagrangian structure yields explicit expressions for kinetic and potential energy.
We exploit this structure to design a model-based proportional-derivative (PD) controller that compensates for the learned potential forces, thereby establishing closed-loop stability guarantees and ensuring convergence.
We evaluate the approach on four simulated DLOs and a real bimanual robot setup.
In simulation, $\Phi$-CON achieves prediction accuracy comparable to neural ordinary differential equation (NODE) models and Lagrangian neural networks (LNNs), with $97.2\%$ and $90.2\%$ fewer parameters, respectively. 
Relative to a model-free PD baseline, the proposed controller reduces mean endpoint regulation error by $97.5$--$99.7\%$ in simulation and achieves a mean per-target reduction of $27.9\%$ across both endpoints in real-world experiments, demonstrating the importance of explicitly accounting for the mechanical response of stiff and heavy objects.\footnote{Video: \href{https://youtu.be/WponGkTDGR8}{YouTube}. Code to be released.}
\end{abstract}

%% file: sections/1-introduction.tex
\section{Introduction} \label{sec:intro}

As robots are increasingly adopted across a wide range of activities, they are required to interact with a more diverse set of objects.
Manipulating \acfp{DLO} such as cables, wires, or tree branches is key to applications like electrical installation, cable routing, and automated harvesting, yet it remains challenging~\citep{billardandkragic2019trendsandchallengesrobotmanip,zhu2022challenges,caporali2026dloroboticsurvey}: \acp{DLO} are high-dimensional, underactuated during manipulation, and often stiff and heavy enough that their physical response cannot be neglected, particularly during rapid motions.
A main body of existing work relies on simplified deformation models of the \ac{DLO}, namely quasi-static assumptions~\citep{bretl2014quasistaticmanipulationkirchhoff} and kinematic descriptions, usually through Jacobian-based deformation estimation~\citep{yu2025generalizablemanipulationDLO} or visual servoing~\citep{shetab2022arapshapeservoing, cao2024shape_control_gnn}, instead of its full dynamics.
Such simplifications are hard to justify for stiff and heavy \acp{DLO}, whose physical response cannot be neglected, especially under rapid motions.

Recent approaches to \ac{DLO} manipulation learn motion models from data using LSTMs~\citep{yang2022biLSTMdynamicmodel}, GNNs~\citep{wang2022cable_gnn, chu2024gladnet}, or recurrent state-space models (RSSMs) \citep{missal2026ropedreamerkinematicrecurrentstate} to predict shape changes from past states and control inputs.
Such architectures can require millions of learned parameters~\citep{missal2026ropedreamerkinematicrecurrentstate}.
Using position or velocity commands as inputs assumes that the manipulator can impose the commanded endpoint motion regardless of \ac{DLO} reaction forces, limiting applicability to lightweight, compliant \acp{DLO}.
Moreover, deploying these learned models for control typically requires numerical optimization at runtime and yields no analytical closed-loop guarantees.

Among learning methods that preserve physical structure~\citep{stolzle2025phdthesis}, \acp{LNN}~\citep{cranmer2020lagrangianneuralnetworks} have been applied to modeling the dynamics of soft robots~\citep{liu2024pinnForSoftRobots} and deformable objects~\citep{zhong2021differentiablecontact, friedl2025reducedlagrangian, pierallini2026cabledynamics}.
However, evaluating their forward dynamics requires runtime differentiation of the learned Lagrangian, including second-order derivatives, which incurs substantial computational cost~\citep{cranmer2020lagrangianneuralnetworks}.
Moreover, Lagrangian structure alone does not guarantee stability of the learned dynamics~\citep{mueller2023exactconservation} or the resulting closed loop~\citep{liu2024pinnForSoftRobots}.
\Acp{CON} combine Lagrangian structure with provably stable latent dynamics and explicit expressions for model-based control~\citep{stolzle2024inputtostateCONmodelbasedcontrolinlatentspace}.
However, their strictly convex latent potential does not accommodate the translational freedom of floating-base \acp{DLO}, while their separately learned observation and input mappings do not enforce consistency between physical and latent forces under the principle of virtual work.

To address these limitations, we propose \textbf{$\mathbf{\Phi}$-CON} for learning force-driven \ac{DLO} dynamics and enabling closed-form control of bimanual manipulation.
We extend \acp{CON} to floating-base systems through a \acf{FBCON}, whose latent potential depends only on shape, accommodating base translation while retaining stable shape dynamics.
A learned affine diffeomorphism $\Phi$ connects observed marker positions to latent coordinates and determines the corresponding force transformation through the principle of virtual work.
The resulting Lagrangian structure enables us to explicitly account for the object's physical response and design a PD model-based controller using the learned potential forces, without requiring runtime optimization.

Our contributions are:
\begin{enumerate}
    \item Closed-form, force-based bimanual regulation of \acp{DLO} through learned potential-force compensation and PD feedback, with closed-loop stability guarantees in observation space for compatible target equilibria of the learned model.
    \item \ac{FBCON}: an extension of \acp{CON} to floating-base systems with explicit kinetic and potential energies and open-loop stability guarantees for the shape dynamics.
    \item $\Phi$-CON: joint learning of \ac{FBCON} dynamics and an affine diffeomorphism that consistently transforms coordinates and forces, preserves Lagrangian structure, and transfers shape-stability guarantees to observation space.
\end{enumerate}
We validate model learning and control using four simulated \acp{DLO} and a real bimanual robotic setup involving the manipulation of a heavy and stiff electrical cable, for which compensating for its physical dynamics is essential.

%% file: sections/2-problem_statement.tex
\section{Problem Statement}\label{sec:problem-statement}

Our goal is to learn the dynamics of a \ac{DLO} from data to enable model-based regulation to a desired shape.
We represent its shape by $N > 2$ marker positions $x_i \in \mathbb{R}^3$, $i = 1, \dots, N$, with actuation only at the two endpoints (\Cref{fig:problem-statement} in the appendix).
Both endpoints are held at a constant orientation.
We order the coordinates as $x = [x_{\mathrm{a}}^\top, x_{\mathrm{u}}^\top]^\top \in \mathbb{R}^{3N}$, where $x_{\mathrm{a}} = [x_1^\top, x_N^\top]^\top \in \mathbb{R}^6$ stacks the actuated endpoints and $x_{\mathrm{u}} = [x_2^\top, \dots, x_{N-1}^\top]^\top \in \mathbb{R}^{3N-6}$ stacks the unactuated interior markers.
The Cartesian endpoint forces $\tau \in \mathbb{R}^6$ induce generalized forces $\tau_x = B\tau$, with $B = [\,I_6 \;\; 0_{6 \times (3N-6)}\,]^\top \in \mathbb{R}^{3N \times 6}$.

\begin{assumption}[Sufficient Observation of DLO Dynamics]\label{ass:backbone-state}
Over the operating regime considered, the $N$ marker positions $x$ are sufficient generalized coordinates for the \ac{DLO}, and $(x,\dot x)$ is a complete dynamical state.
The acceleration $\ddot x$ is uniquely determined by $(x,\dot x,\tau)$ through the Lagrangian dynamics
{\small
\begin{equation}\label{eq:dlo-euler-lagrange-eom}
    M(x) \ddot x + C(x, \dot x) \dot x + D(x) \dot x + \partial_x \mathcal{U}(x) = B \tau,
\end{equation}
}
where $M(x)$ is the mass matrix, $C(x, \dot x)$ is the Coriolis matrix, $D(x)$ a dissipation term, and $\partial_x \mathcal{U}$ the gradient of the potential energy $\mathcal{U}(x)$ collecting both elastic and gravitational contributions.
\end{assumption}

Physically, \Cref{ass:backbone-state} means that the motion is dominated by deformation of the slender backbone, with negligible cross-sectional deformation, as in standard rod models~\citep{alessi2024rodmodelscontinuumsoft}.
It further assumes that independent rotational dynamics, such as twist not captured by the backbone shape, and deformation modes not resolved by the markers have negligible influence on the observed motion.
Under this assumption, we seek to learn the unknown dynamics from sufficiently informative trajectory data $\{x(t_k), \dot x(t_k), \tau(t_k)\}$.

Then, given a desired equilibrium shape $x^d \in \mathbb{R}^{3N}$ supportable by endpoint forces, we seek a model-based feedback law $\tau(x, \dot x)$ such that
{\small
\begin{equation}\label{eq:control-task}
    \lim_{t \to \infty} \|x(t) - x^d\| = 0, \qquad \lim_{t \to \infty} \|\dot x(t)\| = 0.
\end{equation}
}
We assume that $x_{\mathrm{u}}^d$ is a locally unique equilibrium of the unactuated coordinates when the endpoints are fixed at $x_{\mathrm{a}}^d$.
For regulation near this equilibrium, we further assume that $(x_{\mathrm{a}}, \dot x_{\mathrm{a}}) \to (x_{\mathrm{a}}^d, 0)$ implies $(x_{\mathrm{u}}, \dot x_{\mathrm{u}}) \to (x_{\mathrm{u}}^d, 0)$~\citep{pustina2022feebackregularizationunderactuated, borja2022energybasedregulation}.

%% file: sections/3-model_learning.tex
	\section{Dynamical Model Learning with \texorpdfstring{$\Phi$}{Phi}-CON} \label{sec:model-learning}

$\Phi$-CON learns \ac{DLO} dynamics by combining an invertible coordinate mapping with an \ac{FBCON}.
We first separate base position from relative shape, then apply a learned affine map to the shape coordinates.
This representation allows the \ac{FBCON} potential to depend only on shape, accommodating base translation while retaining stable shape dynamics.
The mapping also determines the velocity and force transformations, preserving Lagrangian structure and transferring the stability of the relative shape to observation space.
We train the mapping and latent dynamics jointly using rollout losses in both spaces and a steady-state holding force loss, which encourages accurate predictions of the holding forces used by the controller in \Cref{sec:control}.

\subsection{Translation Invariant Representation}\label{subsec:ti-repr}
\begin{figure}[t]
    \centering
    \includegraphics[width=\linewidth]{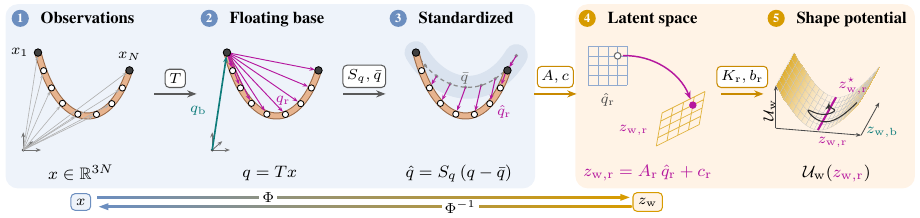}
    \par
    \caption{\textbf{Coordinate transformations of $\Phi$-CON.}
    In \textbf{\textcolor{obsspace}{observation space}} (steps 1-3), fixed maps express the marker positions $x$ as a base $q_{\mathrm{b}}$ and a shape $q_{\mathrm{r}}$ relative to it, and standardize them.
    A learned affine map acting only on the shape leads to \textbf{\textcolor{latspace}{latent space}} (steps 4-5), where the potential is flat along the base, with equilibria $z_{\mathrm{w},\mathrm{r}}^\star$ at any base position.
    }
    \label{fig:phi-con-transformations}
\end{figure}

We choose the first marker as the base, $q_{\mathrm{b}} = x_1 \in \mathbb{R}^3$, and stack the remaining marker positions relative to it as $q_{\mathrm{r}} = [(x_N - x_1)^\top, (x_2 - x_1)^\top, \dots, (x_{N-1} - x_1)^\top]^\top \in \mathbb{R}^{3(N-1)}$, following the ordering of $x$ in \Cref{sec:problem-statement}.
Together, $q = [q_{\mathrm{b}}^\top, q_{\mathrm{r}}^\top]^\top = T x$, where $T$ is a fixed invertible matrix (\Cref{fig:phi-con-transformations}).
A common translation changes only $q_{\mathrm{b}}$, leaving the relative shape $q_{\mathrm{r}}$ unchanged.
We then standardize $q$ as $\hat q = S_q (q - \bar q)$ using the training mean $\bar q$ and a fixed diagonal scaling matrix $S_q$ with strictly positive diagonal entries, preserving invertibility and the separation of base and shape.

\subsection{Affine Diffeomorphism}\label{subsec: affine-difeom}
To complete the mapping into latent space, we apply a learned affine transformation to the relative coordinates while retaining the standardized base:
\begin{equation}\label{eq:affine-mapping-ti-pos}
    z_{\mathrm{w}} = \Phi(x) = A\,\hat q + c,
    \qquad
    A = \operatorname{diag}(I_3, A_{\mathrm{r}}),
    \qquad
    c = [0^\top, c_{\mathrm{r}}^\top]^\top.
\end{equation}
Here, $A_{\mathrm{r}} \in \mathbb{R}^{(3N-3)\times(3N-3)}$ and $c_{\mathrm{r}} \in \mathbb{R}^{3N-3}$ are learned, giving $z_{\mathrm{w},\mathrm{b}} = \hat q_{\mathrm{b}}$ and $z_{\mathrm{w},\mathrm{r}} = A_{\mathrm{r}}\hat q_{\mathrm{r}} + c_{\mathrm{r}}$.
We parameterize $A_{\mathrm{r}} = L_{\mathrm{r}} U_{\mathrm{r}}$ using lower and upper triangular factors with nonzero diagonal entries enforced during training.
Together with the invertibility of $T$ and $S_q$, this makes $\Phi$ an affine diffeomorphism with an exact inverse.

Its Jacobian $J_\Phi = A S_q T$ is independent of $x$ and determines the velocity and force transformations:
\begin{equation}
    \dot z_{\mathrm{w}} = J_\Phi\dot x,
    \qquad
    \tau_{\mathrm{w}} = \alpha J_\Phi^{-\top} B(\tau - \bar\tau).
\end{equation}
The inverse-transpose factor follows from the principle of virtual work.
The mean endpoint force $\bar\tau$ and the positive scale $\alpha$ are computed from the training data and held fixed, with $\alpha$ chosen so that the centered endpoint forces, expressed in the standardized coordinates $\hat q$, have unit root-mean-square value.
The mean force estimates the static load, with $\bar\tau_{\mathrm{L}} + \bar\tau_{\mathrm{R}} \approx -m_{\mathrm{DLO}} g_0$, where $m_{\mathrm{DLO}}$ is the total mass of the \ac{DLO} and $g_0$ is the gravitational acceleration vector.

\subsection{Latent Dynamics: Floating-Base Coupled Oscillator Network}\label{subsec:latent-con}

We extend \ac{CON}~\citep{stolzle2024inputtostateCONmodelbasedcontrolinlatentspace} to floating-base systems by making the latent potential depend only on relative shape:
\begin{equation}\label{eq:latent-w-coords}
    M_{\mathrm{w}} \ddot z_{\mathrm{w}} + D_{\mathrm{w}} \dot z_{\mathrm{w}} + \nabla \mathcal{U}_{\mathrm{w}}(z_{\mathrm{w}}) = \tau_{\mathrm{w}},
\end{equation}
where $M_{\mathrm{w}}$ and $D_{\mathrm{w}}$ are learned, constant, symmetric positive definite mass and damping matrices,\footnote{Positive damping of base translation can represent viscous drag from the surrounding medium, while internal dissipation damps deformation. The coordinate transformation generally couples these contributions, motivating a full positive-definite damping matrix.} and $\tau_{\mathrm{w}}$ is the transformed input from \Cref{subsec: affine-difeom}.
We choose
{\small
\begin{equation}\label{eq:ti-potential}
    \mathcal{U}_{\mathrm{w}}(z_{\mathrm{w}}) = \tfrac12 z_{\mathrm{w},\mathrm{r}}^\top K_{\mathrm{r}} z_{\mathrm{w},\mathrm{r}} + \mathbf{1}^\top\log\cosh(z_{\mathrm{w},\mathrm{r}} + b_{\mathrm{r}}),
    \qquad
    \nabla\mathcal{U}_{\mathrm{w}}(z_{\mathrm{w}}) = \begin{bmatrix} 0 \\ K_{\mathrm{r}} z_{\mathrm{w},\mathrm{r}} + \tanh(z_{\mathrm{w},\mathrm{r}} + b_{\mathrm{r}}) \end{bmatrix},
\end{equation}
}
with learned stiffness $K_{\mathrm{r}} \succ 0$ and bias $b_{\mathrm{r}}$.
This gives translation-invariant dynamics with no potential restoring force along the base, while inertia and damping act on all coordinates.

\begin{proposition}[Stability of Latent FBCON]\label{prop:latent-gas}
For the latent dynamics in Eq.~\eqref{eq:latent-w-coords} with potential energy $\mathcal{U}_{\mathrm{w}}(z_w)$ defined as in Eq.~\eqref{eq:ti-potential}, let $\tau_{\mathrm{w}} = 0$ and $M_{\mathrm{w}}, D_{\mathrm{w}}, K_{\mathrm{r}} \succ 0$.
Then the equilibrium set $\mathcal E_{\mathrm{w}} = \{(z_{\mathrm{w}},\dot z_{\mathrm{w}}): z_{\mathrm{w},\mathrm{r}}=z_{\mathrm{w},\mathrm{r}}^\star,\ \dot z_{\mathrm{w}}=0\}$ is globally asymptotically stable, with arbitrary base position and unique shape $z_{\mathrm{w},\mathrm{r}}^\star$ satisfying
$K_{\mathrm{r}} z_{\mathrm{w},\mathrm{r}}^\star + \tanh(z_{\mathrm{w},\mathrm{r}}^\star + b_{\mathrm{r}}) = 0$.
\end{proposition}

The proof in \Cref{app:proof-latent-gas} adapts the energy argument of~\citet[Thm.~1]{stolzle2024inputtostateCONmodelbasedcontrolinlatentspace} to a potential that depends only on shape.

\subsection{Observation Space Dynamics}\label{subsec:obs-dynamics}

Since $\Phi$ is affine, $\ddot z_{\mathrm{w}} = J_\Phi\ddot x$.
Substituting the coordinate and force transformations into Eq.~\eqref{eq:latent-w-coords} and premultiplying by $\alpha^{-1}J_\Phi^\top$ gives
{\small
\begin{equation}\label{eq:obs-eom}
    M_x\,\ddot x + D_x\,\dot x + g(x) = B\,\tau,
    \qquad
    g(x) := B\bar\tau + \alpha^{-1}J_\Phi^\top\,\nabla\mathcal{U}_{\mathrm{w}}\bigl(\Phi(x)\bigr).
\end{equation}
}%
Here, $M_x = \alpha^{-1}J_\Phi^\top M_{\mathrm{w}} J_\Phi$ and $D_x = \alpha^{-1}J_\Phi^\top D_{\mathrm{w}} J_\Phi$ are constant positive definite matrices.
The dynamics retain Lagrangian structure~\citep[Prop.~2, App.~A]{pustina2024inputdecoupling}, with kinetic energy $\mathcal{T}_x(\dot x)=\tfrac12\dot x^\top M_x\dot x$ and potential energy $\mathcal{U}_x(x) = \alpha^{-1}\mathcal{U}_{\mathrm{w}}(\Phi(x)) + \bar\tau^\top B^\top x$.
Their sum is the total mechanical energy.
The potential gradient $g(x) = \nabla\mathcal{U}_x(x)$ represents the elastic and gravitational load that must be balanced at equilibrium.

\begin{proposition}[Stability of Open-Loop $\Phi$-CON] \label{prop:phi-mapping-gas}
Under the conditions of \Cref{prop:latent-gas}, the dynamics in Eq.~\eqref{eq:obs-eom} with $\tau = \bar\tau$ (equivalently, $\tau_{\mathrm{w}} = 0$) have the globally asymptotically stable equilibrium set $\mathcal E_x = \{(x,\dot x):q_{\mathrm{r}}(x)=q_{\mathrm{r}}^\star,\ \dot x=0\}$, where $q_{\mathrm{r}}^\star$ corresponds uniquely to $z_{\mathrm{w},\mathrm{r}}^\star$ under the affine shape map.
\end{proposition}

The relative shape converges uniquely, while the base converges to a position determined by the initial condition.
Without endpoint forces ($\tau = 0$), the model cannot remain at rest because its estimated gravitational load is uncompensated.
The proof in \Cref{app:proof-open-loop} transfers \Cref{prop:latent-gas} through the invertible affine mapping.

\subsection{Training Strategy} \label{subsec:training-method}

\begin{figure}[ht]
    \centering
    \includegraphics[width=\linewidth]{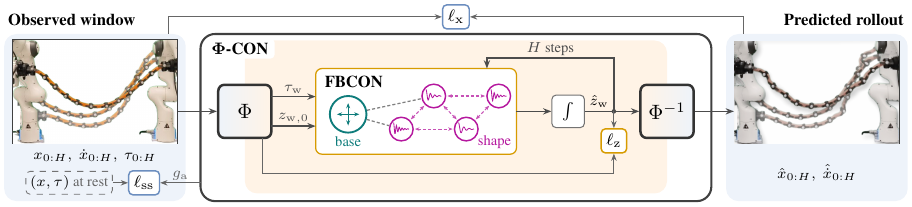}
    \par
    \caption{\textbf{Training of $\Phi$-CON.} $\Phi$ maps observations and inputs from \textbf{\textcolor{obsspace}{observation space}} to \textbf{\textcolor{latspace}{latent space}}, where the \ac{FBCON} is rolled out for $H$ steps and mapped back by $\Phi^{-1}$.
    The losses $\ell_{\mathrm{z}}$ and $\ell_{\mathrm{x}}$ compare the rollout with the data in each space, and $\ell_{\mathrm{ss}}$ fits the holding forces $g_{\mathrm{a}}(x)$ to $\tau$ at rest.
    }
    \label{fig:training-loss}
\end{figure}

We jointly train the coordinate mapping and latent dynamics using $\loss = \loss_{\mathrm{z}} + \loss_{\mathrm{x}} + \lambda_{\mathrm{ss}}\loss_{\mathrm{ss}}$.
\Cref{fig:training-loss} summarizes the training pipeline.
Here, $\loss_{\mathrm{z}}$ and $\loss_{\mathrm{x}}$ measure rollout errors in latent and observation space, respectively, and $\loss_{\mathrm{ss}}$ is the steady-state holding force loss with weight $\lambda_{\mathrm{ss}} > 0$.
$\loss_{\mathrm{x}}$ is the primary objective, while $\loss_{\mathrm{z}}$ acts as a lightly weighted auxiliary term on the latent rollout, and $\loss_{\mathrm{ss}}$ supervises the static forces that rollouts barely constrain.
Exact invertibility of $\Phi$ eliminates the need for a reconstruction loss.
The position and velocity terms in each rollout loss are weighted separately.

For each training trajectory window $\xi = \{(x_t, \dot x_t, \tau_t)\}_{t=0}^{H} \sim \mathcal{D}$, we initialize $\hat z_{\mathrm{w},0} = \Phi(x_0)$ and $\hat{\dot z}_{\mathrm{w},0} = J_\Phi\dot x_0$.
We then predict recursively via $(\hat z_{\mathrm{w},t+1},\hat{\dot z}_{\mathrm{w},t+1}) = f_\theta(\hat z_{\mathrm{w},t},\hat{\dot z}_{\mathrm{w},t},\tau_{\mathrm{w},t})$ for $t = 0,\dots,H-1$, where $f_\theta$ integrates Eq.~\eqref{eq:latent-w-coords} using the transformed input $\tau_{\mathrm{w},t}$.

The \emph{latent rollout} loss $\loss_{\mathrm{z}}$ compares the encoded ground-truth states with the latent rollout:
{\small
\thinmuskip=2mu
\medmuskip=2mu
\thickmuskip=3mu
\begin{equation}
    \loss_{\mathrm{z}} = \mathbb{E}_{\xi \sim \mathcal{D}} \Big[ \tfrac{1}{H+1} \textstyle\sum_{t=0}^{H} \Big( \|\Phi(x_t) - \hat z_{\mathrm{w},t}\|_2^2 + \|J_\Phi \dot x_t - \hat{\dot z}_{\mathrm{w},t}\|_2^2 \Big) \Big].
\end{equation}
}%
The observation-space loss $\loss_{\mathrm{x}}$ compares $x_t$ and $\dot x_t$ with $\Phi^{-1}(\hat z_{\mathrm{w},t})$ and $J_\Phi^{-1}\hat{\dot z}_{\mathrm{w},t}$, respectively, accounting for the effect of the inverse mapping on position and velocity prediction errors.

At equilibrium, Eq.~\eqref{eq:obs-eom} gives $g(x) = B\tau$ and hence $g_{\mathrm{a}}(x) = \tau$, where $g_{\mathrm{a}}(x) := B^\top g(x)$ denotes the potential forces at the actuated endpoints.
We supervise these forces for regulation through $\loss_{\mathrm{ss}} = \tfrac{1}{|\mathcal{D}_{\mathrm{ss}}|} \sum_{(x, \tau) \in \mathcal{D}_{\mathrm{ss}}} \| g_{\mathrm{a}}(x) - \tau \|_2^2$, where $\mathcal{D}_{\mathrm{ss}}$ contains training samples near steady state, selected using the marker velocity and acceleration thresholds in \Cref{app:training-pipeline}.

%% file: sections/4-control.tex
\section{Closed-Loop Control with \texorpdfstring{$\Phi$}{Phi}-CON}\label{sec:control}

Following prior approaches to model-based soft robot control~\citep{borja2022energybasedregulation, dellasantina2023modelbasedcontrolofsoftrobots, stolzle2025phdthesis}, we combine proportional-derivative (PD) feedback with feedforward compensation of the learned potential forces:
{\small
\begin{equation}\label{eq:control-terms}
    \tau = g_{\mathrm{a}}(x^d) + K_{\mathrm{p}} (x_{\mathrm{a}}^d - x_{\mathrm{a}}) - K_{\mathrm{d}}\dot x_{\mathrm{a}}.
\end{equation}
}
The feedforward term $g_{\mathrm{a}}(x^d)$ compensates the learned elastic and gravitational loads at the target.
The PD feedback term adds stiffness and damping in the actuated coordinates to mitigate the effects of modeling errors and external disturbances.

Substituting Eq.~\eqref{eq:control-terms} into the learned dynamics gives
{\small
\begin{equation}\label{eq:cl-underactuated}
    M_x\ddot x + D_x\dot x + g(x) = B \bigl( g_{\mathrm{a}}(x^d) + K_{\mathrm{p}}(x_{\mathrm{a}}^d - x_{\mathrm{a}}) - K_{\mathrm{d}}\dot x_{\mathrm{a}}\bigr).
\end{equation}
}%
Since actuation is restricted to the endpoints, $(x^d, 0)$ is an equilibrium of Eq.~\eqref{eq:cl-underactuated} if and only if $g_{\mathrm{u}}(x^d) = 0$, where $g_{\mathrm{u}}$ collects the unactuated components of $g$.
This condition requires the interior coordinates to be in equilibrium when the endpoints are held at $x_{\mathrm{a}}^d$.

\begin{assumption}\label{ass:gas}
The target satisfies $g_{\mathrm{u}}(x^d) = 0$.
\end{assumption}
\begin{proposition}[Global Asymptotic Stability of Closed-Loop $\Phi$-CON]\label{prop:cl-gas}
Under Assumption~\ref{ass:gas} and the matrix conditions of \Cref{prop:latent-gas}, consider Eq.~\eqref{eq:obs-eom} under the control law in Eq.~\eqref{eq:control-terms}, with gains $K_{\mathrm{p}} \succ 0$ and $K_{\mathrm{d}} \succeq 0$.
Then $(x^d, 0)$ is a globally asymptotically stable equilibrium of the closed-loop dynamics in Eq.~\eqref{eq:cl-underactuated}.
\end{proposition}
The proof in \Cref{app:proof-closed-loop} uses the learned model's mechanical energy modified by the feedforward and proportional terms as a Lyapunov function.
The global guarantee extends to the physical \ac{DLO} if its dynamics match the learned model globally.

%% file: sections/5-simulation.tex
\section{Verification in Simulation}\label{sec:simulation}

\subsection{Simulation Setup}
We evaluate the method in MuJoCo using its discrete elastic rod plugin~\citep{bergou2008discrete-elastic-rods, korner-audoly-bhattacharya2021simple-discrete-elastic-rods}, which excludes extension and shear.
All four simulated \acp{DLO} have length $L = \SI{1}{\meter}$ and $N = 30$ observed markers.
Relative to a baseline \ac{DLO}, the \emph{heavy}, \emph{low damping}, and \emph{compliant} variants change density, damping, and stiffness, respectively; simulator details and physical parameters are given in \Cref{app:sim-params}.

\subsection{Evaluation Metrics and Baseline Methods}

For simulation and real experiments, we split complete trajectories into training and validation sets before extracting rollout windows and evaluate on a separate test set (\Cref{app:dataset-generation}).

We evaluate dynamic prediction, potential-force prediction, and control performance.
For dynamic prediction, we average the position \ac{RMSE} between ground truth $x_t$ and rollout $\hat{x}_t$ over test trajectories: $\mathrm{RMSE}_x(t) = \frac{1}{|\mathcal{D}|}\textstyle\sum_{\xi\in\mathcal{D}} \sqrt{\tfrac{1}{3N}\|\hat{x}_t - x_t\|_2^2}$.
We compute velocity error $\mathrm{RMSE}_{\dot x}(t)$ analogously and report mean $\pm$ standard deviation over seeds.
Following prior work on \ac{DLO} dynamics~\citep{friedl2025reducedlagrangian, pierallini2026cabledynamics}, we compare against \ac{NODE}~\citep{chen2019neuralordinarydifferentialequations} and \ac{LNN}~\citep{cranmer2020lagrangianneuralnetworks}.
The \ac{NODE} predicts $\ddot{x}$ from $(x,\dot{x},\tau_x)$ using a \acf{MLP} without Lagrangian structure.
The \ac{LNN} learns $M(x)$ and $\mathcal{U}(x)$ with separate networks and derives dynamics by automatic differentiation, providing $\partial_x\mathcal{U}$ for potential shaping.
To assess the learned affine diffeomorphism, we also train an \ac{FBCON} (\Cref{subsec:latent-con}) directly in observation space without a learned coordinate mapping.

For control, we report each endpoint's final regulation error, $e_j = \|x_{\mathrm{a},j}(t_{\mathrm{f}}) - x^d_{\mathrm{a},j}\|_2$, as mean $\pm$ standard deviation over targets, pooling both endpoints in simulation.
In real experiments, we also report the per-target relative improvement $(e^{\mathrm{PD}}_j - e^{\mathrm{PD+FF}}_j)/e^{\mathrm{PD}}_j$.
Figures also show the shape error $\tfrac{1}{N}\sum_{i=1}^{N}\|x_i - x_i^d\|_2$, the mean marker distance to the target shape.

\subsection{Dynamic Model Prediction of Simulated DLO}\label{subsec:model-pred-sim}

Within the \SI{1}{s} training horizon, $\Phi$-CON achieves comparable accuracy to \ac{NODE} and \ac{LNN} and outperforms directly employing \ac{FBCON}, supporting the benefit of the learned affine diffeomorphic mapping (\Cref{fig:sim-rollout-node-lnn-phicon}).
Beyond this horizon, \ac{LNN} diverges in the reported rollout comparison.
\ac{LNN} training also exhibits unstable runs and higher computational cost: on the same dataset and an NVIDIA RTX 3090, \ac{NODE}, \ac{LNN}, and $\Phi$-CON require 0.64, 9.35, and 1.06 seconds per epoch, respectively.
Across all four simulated \acp{DLO}, $\Phi$-CON achieves the lowest mean position \ac{RMSE} at \SI{2}{s} (\Cref{tab:rollout-error-all-datasets}).
\Cref{app:rollout-example} illustrates all four models on a training trajectory.

\begin{table}[t]
  \centering
  \caption{\emph{Prediction error in simulation:} Position \ac{RMSE} [mm] for predictions \SI{2}{s} into the future (mean $\pm$ std over seeds).}
  \small
  \setlength{\tabcolsep}{4pt}
  \begin{tabular}{lccc}
    \toprule
    Position RMSE [mm] $\downarrow$ & \ac{NODE} & \ac{LNN} & $\Phi$-CON (ours) \\
    \midrule
    Baseline & $14.6 \pm 0.6$ & $17.6 \pm 1.3$ & $\mathbf{14.0 \pm 0.6}$ \\
    Heavy & $12.5 \pm 1.1$ & $13.6 \pm 0.7$ & $\mathbf{6.6 \pm 0.9}$ \\
    Low damping & $15.4 \pm 1.6$ & $24.8 \pm 1.6$ & $\mathbf{14.8 \pm 1.0}$ \\
    Compliant & $11.0 \pm 1.3$ & $12.9 \pm 0.6$ & $\mathbf{4.5 \pm 0.5}$ \\
    \bottomrule
  \end{tabular}
  \label{tab:rollout-error-all-datasets}
\end{table}

\subsection{Control of the Simulated DLO}\label{subsec:control-dlo-sim}

We evaluate a PD force controller ($K_{\mathrm{p}} = \SI{10}{\newton \per \meter}$, $K_{\mathrm{d}} = \SI{1}{\newton \second \per \meter}$) alone and with potential-force feedforward from an \ac{LNN} or $\Phi$-CON (\Cref{sec:control}).
At test-trajectory steady states, $\Phi$-CON achieves a median relative force-component error of $0.17\%$, compared with $0.13\%$ for \ac{LNN} and $20.47\%$ for an \ac{MLP} trained directly on forces (\Cref{app:force-prediction}).
The models contain $20\,031$, $203\,584$, and $35\,674$ parameters, respectively.

\begin{figure}[tb]
    \centering
    \includegraphics[width=\linewidth]{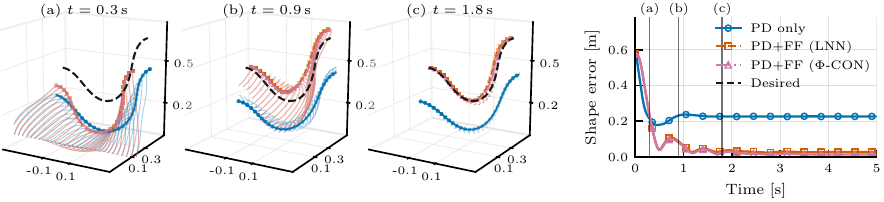}
    \par
    \caption{\emph{Shape regulation in simulation:} regulation from a straight rest configuration to the target shape $x^d$ (\textbf{dashed line}). Left: shape snapshots (axes in meters), with trails showing the preceding \SI{0.6}{s} of motion. Right: shape error over time.
    }
    \label{fig:mujoco-evolution-snapshots}
\end{figure}

Each experiment starts from a straight rest configuration and targets the final steady-state shape $x^d$ of a test trajectory.
PD+FF computes feedforward from the full $x^d$, whereas PD uses only the actuated coordinates $x^d_{\mathrm{a}}$.
Both feedforward models reduce final endpoint errors relative to PD alone and achieve comparable errors across the four datasets (\Cref{tab:error-control-mujoco-all-datasets}).
Low gains keep the closed loop compliant~\citep{dellasantina2017feedbackfeedforward}. Both steady-state errors scale with the inverse of the endpoint stiffness, so the relative reduction barely depends on the gain: on the baseline \ac{DLO} it is $99.4\%$, $99.1\%$, and $98.8\%$ for $K_{\mathrm{p}} = 10$, $50$, and $\SI{200}{\newton\per\meter}$, the gain of the real robots (\Cref{app:gain-sweep}).
\Cref{fig:mujoco-evolution-snapshots} shows an example regulation experiment.

\begin{table}[tb]
  \centering
  \caption{\emph{Regulation error and computational cost in simulation:} Final endpoint regulation error [mm] across the four simulated \ac{DLO} datasets (mean $\pm$ std over test targets and both endpoints), and computation time of one control step for a new target (mean $\pm$ std, \Cref{app:force-prediction}).}
  \small
  \setlength{\tabcolsep}{3pt}

  \begin{tabular}{cccc}
    \toprule
    Endpoint error [mm] $\downarrow$ & PD & PD+FF (\ac{LNN}) & PD+FF ($\Phi$-CON) \\
    \midrule
    Baseline & $238.6 \pm 4.7$ & $2.5 \pm 6.9$ & $\mathbf{1.5 \pm 3.5}$ \\
    Heavy & $459.4 \pm 14.9$ & $\mathbf{2.3 \pm 5.5}$ & $2.7 \pm 5.9$ \\
    Low damping & $238.9 \pm 4.6$ & $\mathbf{3.6 \pm 5.2}$ & $5.9 \pm 9.5$ \\
    Compliant & $240.5 \pm 10.1$ & $\mathbf{0.7 \pm 1.4}$ & $0.8 \pm 0.6$ \\
    \midrule
    CPU time [\si{\micro\second}] $\downarrow$ & $\mathbf{7.8 \pm 1.1}$ & $25.5 \pm 2.9$ & $12.6 \pm 3.4$ \\
    GPU time [\si{\micro\second}] $\downarrow$ & $\mathbf{47.2 \pm 8.9}$ & $114.6 \pm 4.1$ & $72.1 \pm 8.1$ \\
    \bottomrule
  \end{tabular}
  \label{tab:error-control-mujoco-all-datasets}
\end{table}

%% file: sections/6-real_experiments.tex
\section{Validation in a Laboratory Setting}\label{sec:real-experiments}

\subsection{Experimental Setup}\label{subsec:exp-setup}

Two Franka Emika Panda robots firmly grasp the endpoints of a stiff electrical cable of length $L = \SI{0.76}{\metre}$ and radius $r = \SI{9}{\milli\metre}$.
Force sensors measure endpoint forces $\tau$, and motion capture tracks $N = 12$ markers, including the attachment points defining $x_{\mathrm{a}}$.
Identified material parameters are excluded from the modeling and control formulations; characterization, calibration, and preprocessing details are given in \Cref{app:experimental-implementation}.

The robots follow the generated endpoint trajectories from~\Cref{app:dataset-generation} using a bimanual impedance controller~\citep{franzese2024bimanualInteractiveImitationLearning} with stiffness $\SI{200}{\newton \per \meter}$ and damping ratio $0.77$ ($\SI{21.8}{\newton \second \per \meter}$).

\subsection{Dynamic Model Prediction of Real DLO}\label{subsec:model-pred-dlo-real}

\begin{figure}[tb]
    \centering
    \includegraphics[width=0.8\linewidth]{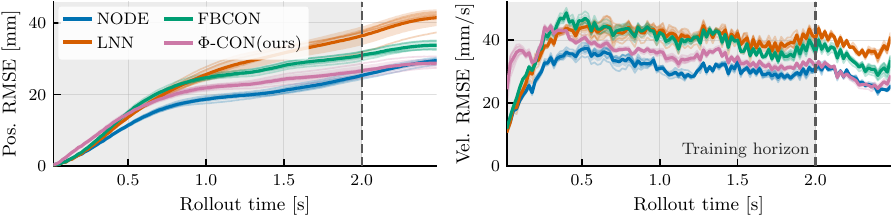}
    \par
    \caption{\emph{Prediction error in experiments:} position (left) and velocity (right) \ac{RMSE} over \SI{2.5}{s} rollouts on the test trajectories of the electrical cable for \textcolor{oiblue}{\textbf{\ac{NODE}}}, \textcolor{oivermillion}{\textbf{\ac{LNN}}}, \textcolor{oigreen}{\textbf{\ac{FBCON}}}, and \textcolor{mplpink}{\textbf{$\mathbf{\Phi}$-CON}}.
    The gray region marks the \SI{2}{s} training horizon.}
    \label{fig:rollout-error-real-electrical-cable}
\end{figure}

\Cref{fig:rollout-error-real-electrical-cable} compares \ac{NODE}, \ac{LNN}, \ac{FBCON}, and $\Phi$-CON rollout errors on recorded test trajectories.
\ac{NODE} is more accurate within the \SI{2}{s} training horizon, and $\Phi$-CON matches it beyond.
\ac{LNN} has the largest errors after approximately \SI{1}{s}.
As in simulation, the learned mapping reduces the errors of the \ac{FBCON}.

\subsection{Control of Real DLO}\label{subsec:control-dlo-real}

\begin{figure*}[t]
    \centering
     \includegraphics[width=1.0\textwidth]{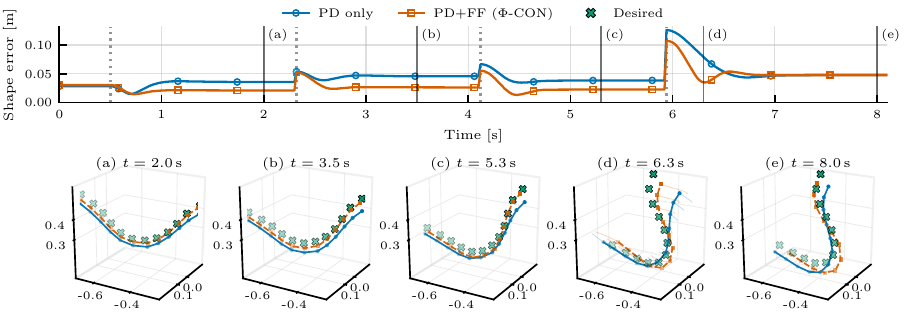}
     \\\vspace{1pt}
     \includegraphics[trim={0 1.5cm 0 0}, clip,width=0.9\textwidth]{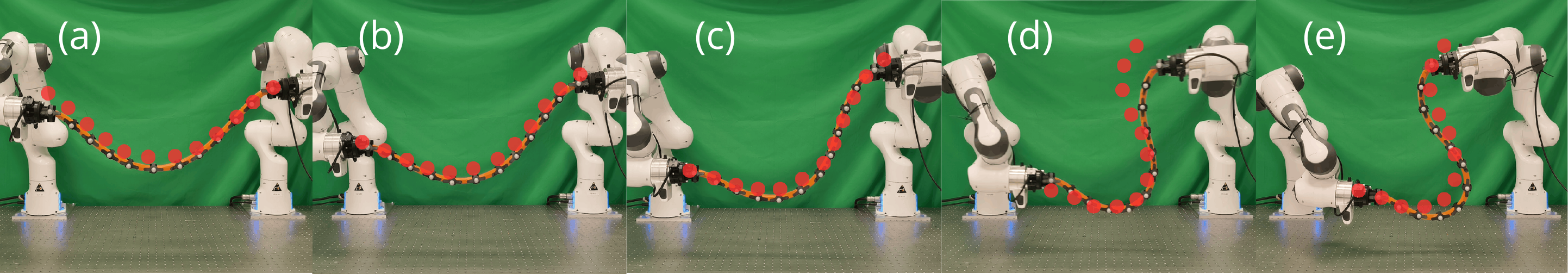}
    \par
    \caption{\emph{Shape regulation in experiments:} regulation to four successive unseen target shapes $x^d$ (changes at dotted lines) with \textcolor{mplblue}{\textbf{PD}} and \textcolor{mplorange}{\textbf{PD+FF ($\Phi$-CON)}}.
    Top: shape error. Middle: targets (\textcolor{mplgreen}{\textbf{green}}) and cable configurations at five instants (axes in meters). Bottom: PD+FF photographs, targets (red).}
    \label{fig:experiment-tracking-plot}
\end{figure*}
We evaluate regulation on 76 steady-state shapes extracted from the training and validation trajectories with the thresholds used for $\mathcal{D}_{\mathrm{ss}}$ (\Cref{app:training-pipeline}), and on four unseen targets obtained by manually moving the robots and recording the cable at rest, which ensures that each target is a feasible equilibrium (\Cref{sec:problem-statement}).
For each target, the feedforward $g_{\mathrm{a}}(x^d)$ in Eq.~\eqref{eq:control-terms} is evaluated once from the recorded $x^d$ and added to the force command of the impedance controller of \Cref{subsec:exp-setup}, which provides the PD term.
Across the steady-state targets, adding $\Phi$-CON feedforward to the PD impedance controller reduces final endpoint errors from $63.7 \pm 8.0$ to $48.4 \pm 5.6$\,mm on the left and from $48.3 \pm 8.5$ to $32.3 \pm 5.2$\,mm on the right.
The corresponding per-target relative improvements are $23.5\% \pm 8.8\%$ and $32.2\% \pm 10.5\%$, respectively.

For the four successive unseen targets in \Cref{fig:experiment-tracking-plot}, PD+FF produces faster transients, particularly for distant targets, and matches or reduces the PD-only steady-state shape error.

%% file: sections/7-conclusion.tex
\section{Conclusions}\label{sec:conclusion}

We presented $\Phi$-CON for learning force-driven \ac{DLO} dynamics by linking an \ac{FBCON} to observed marker positions through an affine diffeomorphism.
Its Lagrangian structure enables PD control with explicit compensation of elastic and gravitational forces, without runtime optimization, and provides closed-loop stability guarantees for feasible equilibria of the learned model.

Across four simulated \acp{DLO}, learned feedforward compensation reduces mean endpoint regulation error by $97.5$--$99.7\%$ relative to PD alone.
On a real cable manipulated by two robots, the learned feedforward reduces the steady-state endpoint error of the robots' impedance controllers by $27.9\%$ per target on average, showing how a learned model can improve low-level manipulator control for \acp{DLO} whose elastic and gravitational loads otherwise cause substantial steady-state errors.
Within the simulation training horizon, $\Phi$-CON achieves prediction accuracy comparable to \ac{NODE} and \ac{LNN}, with $90.2\%$ fewer parameters and $8.8\times$ faster training epochs than \ac{LNN}.

The approach assumes that marker positions and velocities sufficiently describe the \ac{DLO} dynamics (\Cref{ass:backbone-state}) and currently addresses regulation to statically feasible shapes with a fixed grasp orientation, without obstacle interaction or self-contact.
Future work will address material memory, including hysteresis and plastic deformation, through internal states and extend the controller to trajectory tracking.

%% file: sections/acknowledgements.tex
\acks{%
This work was supported in part by the Dutch Research Foundation (NWO) through the VENI grant ROSES 20297, the European Union’s Horizon Europe Programs from Project EMERGE (Grant No. 101070918) and Project RIPLEY (Grant No. 101165078), the Singapore MIT Alliance on Research and Technology (SMART) under the Mens, Manus, et Machina (M3S) program, by the BARI EMERGE program under grant N00014-26-1-2304, and by the MIT-GIST collaboration.

We thank the company ROCSYS for sharing with us the electric cables used in the experimental validations of our method.
The authors also thank Micah Prendergast for sharing the force-torque sensor needed for the validation experiments with the robots.
}

%% file: sections/9-declaration-of-ai.tex
\section*{Declaration of AI Use}

During the preparation of this work, the authors used GPT-6 Astra (OpenAI) and Claude Opus 5.5 (Anthropic) for language editing and proofreading.
Additionally, the authors used Claude Opus 5.5 for code implementation, reproducibility tests, and media formatting, like figures, plots, and editing of the accompanying video.
The authors reviewed the content generated by these tools and remain responsible for it.

%% file: sections/8-appendices.tex
\appendix

\crefalias{section}{appendix}
\crefalias{subsection}{appendix}
\crefname{appendix}{App.}{App.}

\section{Implementation Details}

\begin{figure}[ht]
    \centering
    \includegraphics[width=0.55\linewidth]{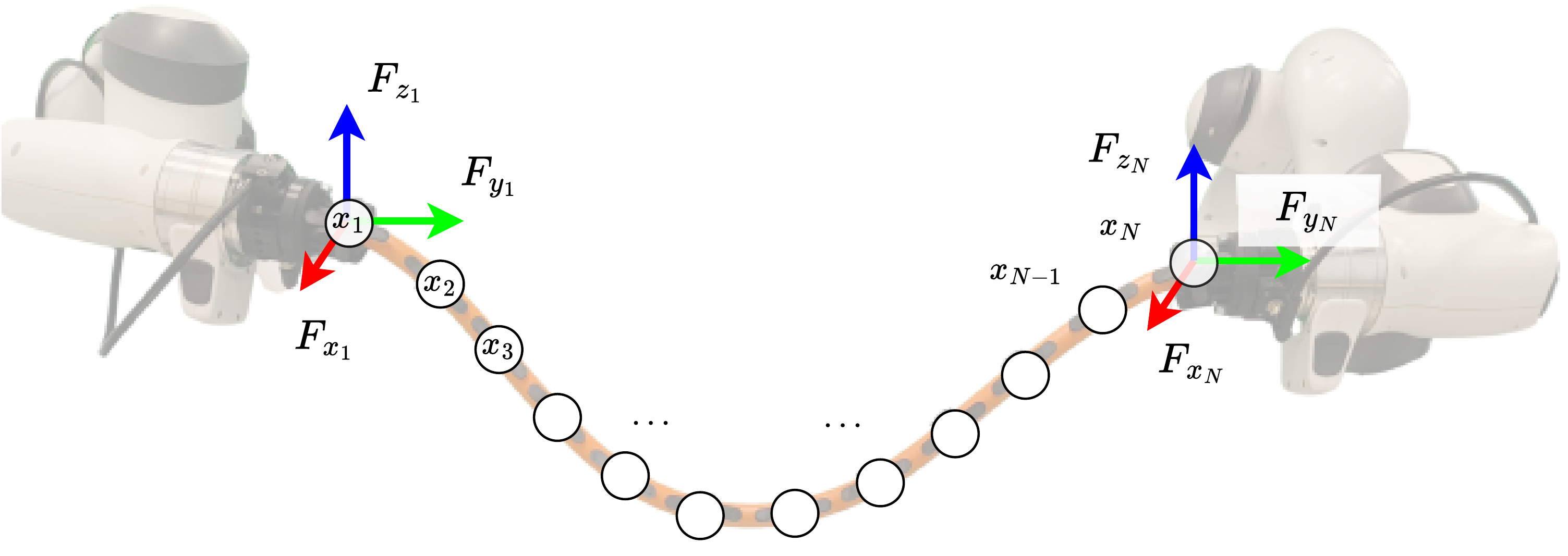}
    \par
    \caption{A \ac{DLO} represented by $N$ 3D marker positions $\{x_1, \dots, x_N\}$ and actuated by Cartesian forces at both endpoints.}
    \label{fig:problem-statement}
\end{figure}

\subsection{Dataset Generation}\label{app:dataset-generation}

We generate endpoint trajectories over a horizon $T = \SI{10}{s}$.
Each endpoint executes one \textbf{linear} or \textbf{oscillatory} motion primitive with randomly sampled class, direction, and amplitude or speed, followed by a \textbf{null} primitive (hold) of random duration, so that the \ac{DLO} reaches a steady state.
In simulation, the hold duration is sampled uniformly between $T_{\mathrm{null}}^{\min} = \SI{0.5}{s}$ and \SI{2}{s}.
We enforce workspace limits and a maximum endpoint separation of $L$, and use PyRoki~\citep{kim2025pyroki} to discard trajectories that are kinematically infeasible for the robots in~\Cref{sec:real-experiments}.
The endpoints track the remaining trajectories using a proportional force controller ($K_{\mathrm{p}} = \SI{50}{\newton\per\metre}$) in MuJoCo.
Each simulation continues for \SI{2}{s} after $T$ with the endpoints held, giving \SI{12}{s} recordings at \SI{1}{kHz}, which we subsample to \SI{100}{Hz} for the dynamics models.

For simulation and real experiments, we randomly split complete trajectories into 80\% training and 20\% validation data before extracting rollout windows.
A separate test set has the same size as the validation set.
All four simulated \acp{DLO} use the same 137 endpoint trajectories for training and validation (109 and 28 per seed).
Their test set contains 28 trajectories generated separately with different random seeds.
The control experiments in \Cref{subsec:control-dlo-sim} target the final steady states of these 28 test trajectories, except on the heavy \ac{DLO}, where two trajectories do not settle (26 targets).
On the real cable, we use 166 recorded trajectories of about \SI{10.5}{s}, after trimming the initial rest to \SI{0.2}{s} before motion onset.
We hold out 28 for testing and use the remaining 138 for training and validation (110 and 28 per seed).

\subsection{Simulated DLO Parameters}\label{app:sim-params}

The MuJoCo discrete elastic rod plugin represents each \ac{DLO} as fixed-length segments connected by ball joints, excluding extension and shear.
Elastic torques resist rotations from the rest configuration, with bending and twisting stiffness set by Young's modulus $E$, shear modulus $G$, and the cross-sectional area moments determined by radius $r$.
Viscous joint damping has coefficient $d$.
All four simulated \acp{DLO} have length $L = \SI{1}{\meter}$, $N = 30$ observed markers, and radius $r = \SI{0.005}{\meter}$.
\Cref{tab:cable_params_all} lists the baseline parameters and the variants, which change density (\emph{heavy}), damping (\emph{low damping}), or stiffness (\emph{compliant}); all other parameters remain at baseline values.

\begin{table}[ht]
    \centering
    \caption{Physical parameters of the four simulated \acp{DLO}. Bold values differ from the baseline.}
    \begin{tabular}{lcccc}
        \toprule
        \ac{DLO} Dataset & $E$ [Pa] & $G$ [Pa] & $\rho$ [kg/m$^3$] & $d$ [N$\cdot$m$\cdot$s/rad] \\
        \midrule
        Baseline & $6{\times}10^{7}$ & $1{\times}10^{7}$ & $3{\times}10^{3}$ & $1.5{\times}10^{-2}$ \\
        Heavy    & $6{\times}10^{7}$ & $1{\times}10^{7}$ & $\mathbf{7.8{\times}10^{3}}$ & $1.5{\times}10^{-2}$ \\
        Low damping & $6{\times}10^{7}$ & $1{\times}10^{7}$ & $3{\times}10^{3}$ & $\mathbf{3{\times}10^{-3}}$ \\
        Compliant   & $\mathbf{3{\times}10^{6}}$ & $\mathbf{1{\times}10^{6}}$ & $3{\times}10^{3}$ & $1.5{\times}10^{-2}$ \\
        \bottomrule
    \end{tabular}
    \label{tab:cable_params_all}
\end{table}

\subsection{Training Pipeline}\label{app:training-pipeline}

For each model we train five seeds, each setting the initialization and the train/validation split, and keep the checkpoint with the lowest validation loss.

For the steady-state holding force loss, $\mathcal{D}_{\mathrm{ss}}$ contains training samples $(x_t, \tau_t)$ for which every marker satisfies $\|\dot x_{i,t}\| < \SI{0.01}{m/s}$ and $\|\ddot x_{i,t}\| < \SI{0.1}{m/s^2}$.

\subsection{Hyperparameters}
All models are trained with AdamW (constant learning rate, gradient-norm clipping at 1) on batches of 512 windows of 100 steps, integrated with explicit Euler at the data rate: $\Delta t=\SI{10}{ms}$ (\SI{1}{s} windows) in simulation and $\Delta t=\SI{20}{ms}$ (\SI{2}{s} windows) on the real cable.
States and forces are standardized with training-set statistics (\Cref{subsec:ti-repr}), and $\bar\tau$ and $\alpha$ are computed from the training data and kept fixed (\Cref{subsec: affine-difeom}).
Hyperparameters are tuned on the baseline simulated DLO and reused unchanged for the other three DLOs, and tuned separately for the real cable.
\Cref{tab:hyperparams} lists the optimization and loss hyperparameters of all models, \Cref{tab:hparams-node-lnn} the network architectures of the \ac{NODE} and \ac{LNN}, and \Cref{tab:hparams-con} the latent-space settings of the \ac{FBCON} and $\Phi$-CON.
The \ac{NODE} and \ac{LNN} receive the base position as a zero input, which makes them translation invariant like the \ac{FBCON} and $\Phi$-CON.
For the \ac{FBCON}, the learned map of $\Phi$ is fixed to the identity (i.e., $A = I$, $c = 0$), so $\loss_z$ compares standardized coordinates.

The \ac{LNN} is conservative, i.e., it has no dissipation term, as in~\citet{cranmer2020lagrangianneuralnetworks}.
We also trained a damped variant following Eq.~\eqref{eq:dlo-euler-lagrange-eom} with a learned constant $D \succ 0$, parameterized and initialized like the damping of $\Phi$-CON, but its training diverged.

We tune each model with Optuna's TPE sampler over 40--60 trials of 1000 epochs (500 for the \ac{LNN}), pruning diverging and dominated trials early.
The search includes the rollout loss weights for position and velocity and the steady-state holding force loss weight $\lambda_{\mathrm{ss}}$.
Trials are scored by the validation rollout error in observation space, using fixed position and velocity weights so that trials with different loss weights remain comparable, and, for models with a steady-state holding force loss (\ac{LNN}, \ac{FBCON}, $\Phi$-CON), by the median validation endpoint-force error.
We select a Pareto-optimal trial; the search ranges are listed in \Cref{tab:search-space}.

\begin{table}[ht]
  \centering
  \caption{Optimization and loss hyperparameters of all models. $w_x^{\mathrm{vel}}$: weight of the velocity term in $\loss_x$ (position term weighted 1); ---: not part of the model.}
  \label{tab:hyperparams}
  \small
  \begin{tabular}{lcccc}
    \toprule
    & \ac{NODE} & \ac{LNN} & \ac{FBCON} & $\Phi$-CON \\
    \midrule
    \multicolumn{5}{l}{\emph{Simulation (all four DLOs)}} \\
    Learning rate & \num{6.5e-4} & \num{1.0e-4}\,$^{\dagger}$ & \num{2.9e-3} & \num{2.2e-3} \\
    Weight decay & 0 & 0 & \num{3.9e-4} & \num{3.8e-2} \\
    $w_x^{\mathrm{vel}}$ & 1.27 & 29.0 & 13.1 & 31.3 \\
    $\lambda_{\mathrm{ss}}$ & --- & 0.36 & 0.65 & 0.50 \\
    Epochs & 5000 & 2000 & 5000 & 5000 \\
    Parameters & \num{710234} & \num{203584} & \num{12375} & \num{20031} \\
    \midrule
    \multicolumn{5}{l}{\emph{Real electrical cable}} \\
    Learning rate & \num{1.1e-3} & \num{2.0e-4}\,$^{\dagger}$ & \num{1.1e-3} & \num{4.3e-3} \\
    Weight decay & 0 & 0 & \num{5.5e-4} & \num{2.4e-4} \\
    $w_x^{\mathrm{vel}}$ & 0.53 & 26.9 & 6.09 & 1.04 \\
    $\lambda_{\mathrm{ss}}$ & --- & \num{1.8e-3} & 0.18 & \num{4.2e-4} \\
    Epochs & 2000 & 2000 & 5000 & 5000 \\
    Parameters & \num{336932} & \num{200603} & \num{2034} & \num{3156} \\
    \bottomrule
  \end{tabular}
  \par\smallskip\footnotesize
  $^{\dagger}$Lowered from the tuned value (\num{1.2e-4} and \num{2.5e-4}) for training stability.
\end{table}

\begin{table}[ht]
  \centering
  \begin{minipage}[t]{0.48\linewidth}
    \centering
    \caption{Hidden layers $\times$ width of the \ac{NODE} and of the \ac{LNN} mass network $M$ and potential network $\mathcal{U}$ (swish activations).}
    \label{tab:hparams-node-lnn}
    \small
    \setlength{\tabcolsep}{4pt}
    \begin{tabular}{lccc}
      \toprule
      & \ac{NODE} & \ac{LNN} $M$ & \ac{LNN} $\mathcal{U}$ \\
      \midrule
      Simulation & $3\times512$ & $5\times32$ & $4\times128$ \\
      Real cable & $2\times512$ & $6\times128$ & $7\times64$ \\
      \bottomrule
    \end{tabular}
  \end{minipage}\hfill
  \begin{minipage}[t]{0.48\linewidth}
    \centering
    \caption{Mapping $\Phi$ and weights $w_z^{\mathrm{pos}}$, $w_z^{\mathrm{vel}}$ of the position and velocity terms in $\loss_z$.}
    \label{tab:hparams-con}
    \small
    \setlength{\tabcolsep}{4pt}
    \begin{tabular}{lcc}
      \toprule
      & \ac{FBCON} & $\Phi$-CON \\
      \midrule
      Mapping $\Phi$ & identity & affine \\
      \addlinespace
      \multicolumn{3}{l}{\emph{Simulation}} \\
      $w_z^{\mathrm{pos}}$ & \num{2.2e-3} & \num{5.6e-5} \\
      $w_z^{\mathrm{vel}}$ & \num{1.6e-5} & \num{5.8e-4} \\
      \addlinespace
      \multicolumn{3}{l}{\emph{Real cable}} \\
      $w_z^{\mathrm{pos}}$ & \num{1.9e-3} & \num{1.3e-5} \\
      $w_z^{\mathrm{vel}}$ & \num{3.2e-4} & \num{1.5e-3} \\
      \bottomrule
    \end{tabular}
  \end{minipage}
\end{table}

\begin{table}[ht]
  \centering
  \caption{Hyperparameter search ranges (log-uniform unless stated).}
  \label{tab:search-space}
  \small
  \begin{tabular}{lll}
    \toprule
    Hyperparameter & Range & Models \\
    \midrule
    Learning rate & $[10^{-6}, 10^{-2}]$ / $[10^{-5}, 10^{-2}]$ / $[10^{-5}, \num{3e-3}]$ & \ac{NODE}, \ac{LNN} / $\Phi$-CON / \ac{FBCON} \\
    Weight decay & $\{0\}\cup[10^{-4}, 10^{-2}]$ / $[10^{-4}, 10^{-1}]$ & \ac{NODE}, \ac{LNN}, \ac{FBCON} / $\Phi$-CON \\
    $w_x^{\mathrm{vel}}$ & $[10^{-2}, 10^{2}]$ & all \\
    $w_z^{\mathrm{pos}}$, $w_z^{\mathrm{vel}}$ & $[10^{-5}, 10^{2}]$ & $\Phi$-CON, \ac{FBCON} (optional) \\
    $\lambda_{\mathrm{ss}}$ & $[10^{-4}, 1]$ & \ac{LNN}, \ac{FBCON}, $\Phi$-CON \\
    Hidden layers & $\{2,\dots,8\}$ & \ac{NODE}; \ac{LNN} (each network) \\
    Width & $\{64,128,256,512\}$ / $\{32,64,128\}$ & \ac{NODE} / \ac{LNN} \\
    \bottomrule
  \end{tabular}
\end{table}

\subsection{Experimental Implementation}\label{app:experimental-implementation}

The electrical cable has identified density $\rho \approx \SI{3.5e3}{\kilogram\per\cubic\metre}$, Young's modulus $E \approx \SI{22}{\mega\pascal}$, and elastic damping coefficient $\nu \approx \SI{2}{\mega\pascal\second}$.
These values characterize the material and are excluded from the modeling and control formulations.
Two Bota Systems SenseOne sensors measure endpoint forces $\tau$, with calibration before recording and drift correction using a per-trajectory offset.
Five OptiTrack cameras track $N = 12$ markers, including the attachment points defining $x_{\mathrm{a}}$.
Each trajectory starts from the same robot joint configuration to limit variations in force capability, since robot dynamics are not modeled.
To reduce noise amplification, we downsample marker positions from $\SI{240}{\hertz}$ to $\SI{50}{\hertz}$ before computing velocities by central finite differences.

\section{Dynamic Prediction in Simulation}\label{app:rollout-example}

\begin{figure}[tb]
    \centering
    \includegraphics[width=\linewidth]{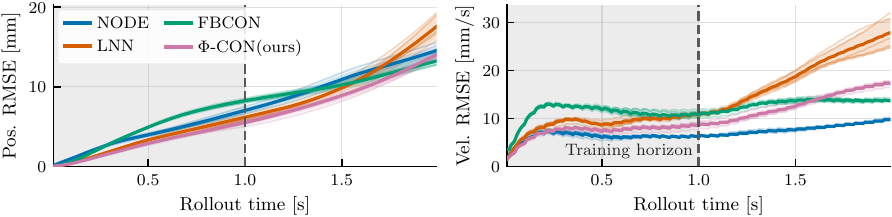}
    \par\vspace{-8pt} %
    \caption{\emph{Prediction error in simulation:} position (left) and velocity (right) \ac{RMSE} over \SI{2}{s} rollouts for \textcolor{oiblue}{\textbf{\ac{NODE}}}, \textcolor{oivermillion}{\textbf{\ac{LNN}}}, \textcolor{oigreen}{\textbf{\ac{FBCON}}}, and \textcolor{mplpink}{\textbf{$\mathbf{\Phi}$-CON}} on the baseline dataset.
    The gray region marks the \SI{1}{s} training horizon.
    }
    \label{fig:sim-rollout-node-lnn-phicon}
\end{figure}

\begin{figure}[ht]
    \centering
    \includegraphics[width=\linewidth]{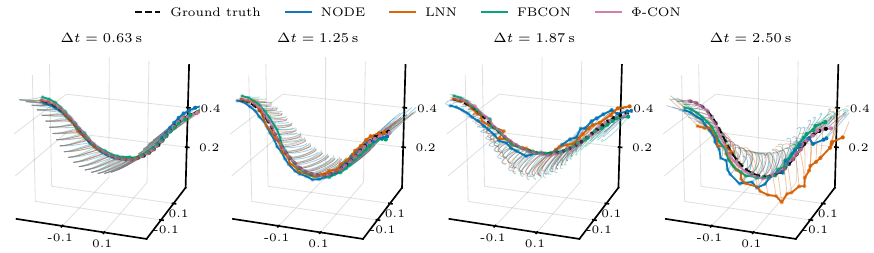}
    \par\vspace{-8pt} %
    \caption{\emph{Motion prediction in simulation:} predictions along a high-velocity trajectory from \textcolor{mplblue}{\textbf{\ac{NODE}}}, \textcolor{mplorange}{\textbf{\ac{LNN}}}, \textcolor{mplgreen}{\textbf{\ac{FBCON}}}, and \textcolor{mplpink}{\textbf{$\mathbf{\Phi}$-CON}}.
    Thin lines trace motion between snapshots.
    \ac{NODE} accumulates error, \ac{LNN} diverges beyond its \SI{1}{s} training horizon, and \ac{FBCON} lags behind the ground truth; $\Phi$-CON follows the motion throughout.}
    \label{fig:evolution-rollout-comparison}
\end{figure}

\Cref{fig:evolution-rollout-comparison} shows a rollout of all four models along a training trajectory, complementing the baseline position and velocity prediction errors in \Cref{fig:sim-rollout-node-lnn-phicon} and the discussion in \Cref{subsec:model-pred-sim}.

\section{Prediction of Potential Forces for Simulated DLO}\label{app:force-prediction}

We evaluate potential-force prediction, which provides the feedforward term in our controller, against an \ac{MLP} trained directly on forces, $\hat \tau_x = \operatorname{MLP}(x, \dot x, \ddot x)$, and an \ac{LNN}~\citep{cranmer2020lagrangianneuralnetworks}.
At steady state, the \ac{MLP} predicts potential forces as $\hat \tau_{x_{\mathrm{ss}}} = \operatorname{MLP}(x_{\mathrm{ss}}, 0, 0)$.
The \ac{MLP} has four hidden layers of width 64 with ReLU activations, and its inputs are expressed relative to the first marker with that marker's absolute position removed, so that, like the other models, it is translation invariant.
It is trained on the baseline training trajectories, sampled at \SI{500}{Hz}, with a mean-squared force loss over all time steps, using AdamW (learning rate and weight decay $10^{-4}$, gradient-norm clipping at 1) on batches of 256 windows for 2000 epochs.
We train three seeds and keep, for each, the checkpoint with the lowest validation loss.
For each test trajectory, we extract the final steady state $x^i_{\mathrm{ss}}$ and endpoint force $\tau^i_{\mathrm{ss}}\in\mathbb{R}^6$, and compute the relative error $e^{i}_{j,k} = \big|\hat g_{\mathrm{a},j,k}(x^i_{\mathrm{ss}}) - \tau^i_{\mathrm{ss},j,k}\big| \,/\, \|\tau^i_{\mathrm{ss},j}\|_2$, for endpoint $j\in\{\text{left},\text{right}\}$ and axis $k\in\{X,Y,Z\}$.
\Cref{tab:force-prediction-error} reports the median over trajectories for all three models.
$\Phi$-CON outperforms the \ac{MLP} and matches the \ac{LNN} with fewer parameters: $20\,031$, compared with $35\,674$ for the \ac{MLP} and $203\,584$ for the \ac{LNN}.
We also report the computation time of one prediction for a single state, measured with JAX on an AMD Ryzen~9 5900HS CPU and an NVIDIA RTX~3050~Ti Laptop GPU as mean $\pm$ std over $10^4$ calls of the compiled function.
For $\Phi$-CON, $K_{\mathrm{r}}$ and $A_r = L_r U_r$ depend only on the learned parameters, so we assemble them once after training instead of at every call, and $g(x)$ reduces to a few matrix-vector products.
For a single state, the GPU is slower than the CPU because kernel-launch overhead dominates the computation.
The control times in \Cref{tab:error-control-mujoco-all-datasets} are measured in the same way, for one PD or PD+FF step with a new target $x^d$ compiled as a single function.

\begin{table}[ht]
  \centering
  \caption{\emph{Force prediction error and computational cost in simulation:} Median relative endpoint potential-force prediction error, normalized by the endpoint force magnitude, computation time per prediction (mean $\pm$ std), and number of parameters.}
  \label{tab:force-prediction-error}
  \small
  \setlength{\tabcolsep}{2pt}
  \begin{tabular}{lcccccccccc}
    \toprule
    & \multicolumn{3}{c}{Left endpoint} & \multicolumn{3}{c}{Right endpoint} & & \multicolumn{2}{c}{Time [\si{\micro\second}]} & \\
    \cmidrule(lr){2-4} \cmidrule(lr){5-7} \cmidrule(lr){9-10}
    Model & X & Y & Z & X & Y & Z & All & CPU & GPU & Params \\
    \midrule
    \ac{MLP} & $39.77\%$ & $12.90\%$ & $16.86\%$ & $40.78\%$ & $6.55\%$ & $24.89\%$ & $20.47\%$ & $15.6 \pm 2.4$ & $66.5 \pm 7.6$ & $35\,674$ \\
    \ac{LNN} & $\mathbf{0.12\%}$ & $0.23\%$ & $\mathbf{0.09\%}$ & $\mathbf{0.11\%}$ & $0.27\%$ & $\mathbf{0.08\%}$ & $\mathbf{0.13\%}$ & $25.4 \pm 3.9$ & $105.2 \pm 5.0$ & $203\,584$ \\
    $\Phi$-CON & $0.19\%$ & $\mathbf{0.20\%}$ & $0.15\%$ & $0.20\%$ & $\mathbf{0.20\%}$ & $0.14\%$ & $0.17\%$ & $\mathbf{13.9 \pm 3.0}$ & $\mathbf{64.8 \pm 15.8}$ & $\mathbf{20\,031}$ \\
    \bottomrule
  \end{tabular}
\end{table}

\section{Gain Sensitivity of Simulated Regulation}\label{app:gain-sweep}

We repeat the baseline regulation experiment of \Cref{subsec:control-dlo-sim} with higher PD gains, scaling $K_{\mathrm{d}}$ with $\sqrt{K_{\mathrm{p}}}$ to keep the damping ratio, while the model, targets, and settings are unchanged (\Cref{tab:gain-sweep}).
The PD error scales with $1/K_{\mathrm{p}}$, since PD alone never supplies the holding force.
The PD+FF error also decreases with the gain, so the relative reduction stays between $98.8\%$ and $99.4\%$, including at the \SI{200}{\newton\per\meter} used on the real robots.

\begin{table}[ht]
  \centering
  \caption{\emph{Regulation error in simulation for varying control gains:} Final endpoint regulation error [mm] on the baseline \ac{DLO} for increasing PD gains, with $K_{\mathrm{d}} \propto \sqrt{K_{\mathrm{p}}}$ to keep the damping ratio (mean $\pm$ std over the 28 test targets and both endpoints).}
  \label{tab:gain-sweep}
  \begin{tabular}{ccccc}
    \toprule
    $K_{\mathrm{p}}$ [N/m] & $K_{\mathrm{d}}$ [N\,s/m] & PD & PD+FF ($\Phi$-CON) & Reduction \\
    \midrule
    10  & 1.0  & $238.6 \pm 4.7$ & $1.5 \pm 3.5$   & $99.4\%$ \\
    50  & 2.24 & $48.2 \pm 1.1$  & $0.44 \pm 0.66$ & $99.1\%$ \\
    200 & 4.47 & $12.1 \pm 0.4$  & $0.15 \pm 0.21$ & $98.8\%$ \\
    \bottomrule
  \end{tabular}
\end{table}

\section{Proofs of Stability}\label{app:proofs-stability}

\subsection{Proof of Proposition~\ref{prop:latent-gas}: Stability of Latent FBCON}\label{app:proof-latent-gas}
As in~\citet[Thm.~1]{stolzle2024inputtostateCONmodelbasedcontrolinlatentspace}, choose the mechanical energy
\begin{equation}\label{eq:latent-lyapunov}
    V = \tfrac12\dot z_{\mathrm{w}}^\top M_{\mathrm{w}}\dot z_{\mathrm{w}} + \mathcal{U}_{\mathrm{w}}(z_{\mathrm{w}}) - \mathcal{U}_{\mathrm{w}}(z_{\mathrm{w}}^\star),
\end{equation}
where $(z_{\mathrm{w}}^\star,0)\in\mathcal E_{\mathrm{w}}$.
Since $M_{\mathrm{w}}$ is constant, differentiating $V$ and substituting Eq.~\eqref{eq:latent-w-coords} with $\tau_{\mathrm{w}}=0$ gives
\begin{equation}\label{eq:latent-lyapunov-derivative}
\begin{aligned}
    \dot V
    &=\dot z_{\mathrm{w}}^\top M_{\mathrm{w}}\ddot z_{\mathrm{w}}
      +\nabla\mathcal U_{\mathrm{w}}^\top\dot z_{\mathrm{w}}\\
    &=\dot z_{\mathrm{w}}^\top\bigl(M_{\mathrm{w}}\ddot z_{\mathrm{w}}+\nabla\mathcal U_{\mathrm{w}}\bigr)\\
    &=-\dot z_{\mathrm{w}}^\top D_{\mathrm{w}}\dot z_{\mathrm{w}}\le0.
\end{aligned}
\end{equation}
The shape Hessian is
\begin{equation}\label{eq:latent-shape-hessian}
    \nabla^2_{z_{\mathrm{w},\mathrm{r}}}\mathcal U_{\mathrm{w}}
    =K_{\mathrm{r}}+\operatorname{diag}\!\bigl(\operatorname{sech}^2(z_{\mathrm{w},\mathrm{r}}+b_{\mathrm{r}})\bigr)
    \succeq K_{\mathrm{r}}\succ0,
\end{equation}
so $\mathcal U_{\mathrm{w}}$ is strongly convex in the shape, while it is independent of the base $z_{\mathrm{w},\mathrm{b}}$.
Since the dynamics do not depend on $z_{\mathrm{w},\mathrm{b}}$ either, $(z_{\mathrm{w},\mathrm{r}}, \dot z_{\mathrm{w}})$ evolves as an autonomous subsystem.
Strong convexity and $M_{\mathrm{w}}\succ0$ make $V$ positive definite about $(z_{\mathrm{w},\mathrm{r}}^\star,0)$ and radially unbounded in this reduced state, giving compact positively invariant sublevel sets and Lyapunov stability.
Since $D_{\mathrm{w}}\succ0$, $\dot V=0$ requires $\dot z_{\mathrm{w}}=0$; remaining there requires $\nabla\mathcal U_{\mathrm{w}}=0$, hence $z_{\mathrm{w},\mathrm{r}}=z_{\mathrm{w},\mathrm{r}}^\star$.
LaSalle's invariance principle therefore gives global asymptotic stability of the reduced equilibrium and, equivalently, of $\mathcal E_{\mathrm{w}}$, whose Euclidean distance is $\bigl(\|z_{\mathrm{w},\mathrm{r}}-z_{\mathrm{w},\mathrm{r}}^\star\|^2+\|\dot z_{\mathrm{w}}\|^2\bigr)^{1/2}$.

\begin{samepage}
The base also converges to a finite position.
Partitioning $M_{\mathrm{w}}$ and $D_{\mathrm{w}}$ into base and shape blocks, the base equation is
\begin{equation}\label{eq:latent-base-dynamics}
    M_{\mathrm{w},\mathrm{bb}}\ddot z_{\mathrm{w},\mathrm{b}}
    +M_{\mathrm{w},\mathrm{br}}\ddot z_{\mathrm{w},\mathrm{r}}
    +D_{\mathrm{w},\mathrm{bb}}\dot z_{\mathrm{w},\mathrm{b}}
    +D_{\mathrm{w},\mathrm{br}}\dot z_{\mathrm{w},\mathrm{r}}=0.
\end{equation}
Integrating with respect to time gives
\begin{equation}\label{eq:latent-base-integral}
    M_{\mathrm{w},\mathrm{bb}}\dot z_{\mathrm{w},\mathrm{b}}
    +M_{\mathrm{w},\mathrm{br}}\dot z_{\mathrm{w},\mathrm{r}}
    +D_{\mathrm{w},\mathrm{bb}}z_{\mathrm{w},\mathrm{b}}
    +D_{\mathrm{w},\mathrm{br}}z_{\mathrm{w},\mathrm{r}}=c_{\mathrm{b}},
\end{equation}
where $c_{\mathrm{b}}$ is an integration constant determined by the initial state, encoding the balance between generalized base momentum and accumulated damping forces.
As $\dot z_{\mathrm{w}}\to0$ and $z_{\mathrm{w},\mathrm{r}}\to z_{\mathrm{w},\mathrm{r}}^\star$, invertibility of $D_{\mathrm{w},\mathrm{bb}}\succ0$ yields
\begin{equation}\label{eq:latent-base-limit}
    z_{\mathrm{w},\mathrm{b}}\to D_{\mathrm{w},\mathrm{bb}}^{-1}
    \bigl(c_{\mathrm{b}}-D_{\mathrm{w},\mathrm{br}}z_{\mathrm{w},\mathrm{r}}^\star\bigr).
\end{equation}
\end{samepage}

\subsection{Proof of Proposition~\ref{prop:phi-mapping-gas}: Stability of Open-Loop \texorpdfstring{$\Phi$}{Phi}-CON}\label{app:proof-open-loop}
The affine bijection $\Phi$ preserves the separation of base and shape: $z_{\mathrm{w},\mathrm{r}}$ is an invertible affine function of $q_{\mathrm{r}}$, and $\dot x = J_\Phi^{-1}\dot z_{\mathrm{w}}$.
Under $\tau=\bar\tau$, the constant input compensates the linear load term $\bar\tau^\top B^\top x$ in $\mathcal U_x$.
Expressing the Lyapunov function from \Cref{app:proof-latent-gas} in observation coordinates gives
\begin{equation}\label{eq:obs-open-loop-lyapunov}
    V_x(x,\dot x)=\alpha^{-1}V\bigl(\Phi(x),J_\Phi\dot x\bigr).
\end{equation}
Its time derivative satisfies
\begin{equation}\label{eq:obs-open-loop-lyapunov-derivative}
    \dot V_x=-\alpha^{-1}\dot x^\top J_\Phi^\top D_{\mathrm{w}}J_\Phi\dot x
              =-\dot x^\top D_x\dot x\le0.
\end{equation}
The induced invertible linear map between reduced errors preserves both stability and convergence, proving global asymptotic stability of $\mathcal E_x$.
Base-position convergence also transfers through the fixed affine standardization.

\subsection{Proof of Proposition~\ref{prop:cl-gas}: Global Asymptotic Stability of Closed-Loop System}\label{app:proof-closed-loop}
The proof uses the Lyapunov function
{\small
\begin{equation}\label{eq:cl-lyapunov}
    V = \tfrac12\dot x^\top M_x\dot x + \mathcal{U}_x(x) - \mathcal{U}_x(x^d) - g(x^d)^\top(x - x^d) + \tfrac12(x_{\mathrm{a}} - x_{\mathrm{a}}^d)^\top K_{\mathrm{p}}(x_{\mathrm{a}} - x_{\mathrm{a}}^d),
\end{equation}
}
where $\mathcal{U}_x$ is the potential of $g$ (\Cref{subsec:obs-dynamics}).
Let $W(x)$ denote the position part of $V$.
Its gradient is
\begin{equation}\label{eq:cl-potential-gradient}
    \nabla W(x)=g(x)-g(x^d)+BK_{\mathrm{p}}(x_{\mathrm{a}}-x_{\mathrm{a}}^d),
\end{equation}
which vanishes at $x=x^d$.
Using the affine map in \Cref{subsec: affine-difeom},
\begin{equation}\label{eq:cl-potential-hessian}
    \nabla^2 W(x)\succeq \alpha^{-1}T^\top S_q^\top
    \operatorname{diag}(0_{3\times3},A_{\mathrm{r}}^\top K_{\mathrm{r}}A_{\mathrm{r}})
    S_q T+BK_{\mathrm{p}}B^\top\succ0.
\end{equation}
The first term penalizes every displacement that changes the relative shape.
Any other nonzero displacement translates all markers equally, moving the endpoints and thus being penalized by the second term.
The constant positive-definite lower bound makes $W$ strongly convex, with its unique minimum $W(x^d)=0$.
Together with $M_x\succ0$, this makes $V$ positive definite about $(x^d,0)$ and radially unbounded.
Feasibility gives $g(x^d)=Bg_{\mathrm{a}}(x^d)$, so the closed-loop dynamics are $M_x\ddot x+(D_x+BK_{\mathrm{d}}B^\top)\dot x+\nabla W(x)=0$.
Differentiating $V$ and using Eq.~\eqref{eq:obs-eom} and feasibility in the second line, then substituting the controller in the third, yields
\begin{equation}\label{eq:cl-lyapunov-derivative}
\begin{aligned}
    \dot V
    &=\dot x^\top\bigl(M_x\ddot x+g(x)-g(x^d)\bigr)
      +\dot x_{\mathrm{a}}^\top K_{\mathrm{p}}(x_{\mathrm{a}}-x_{\mathrm{a}}^d)\\
    &=-\dot x^\top D_x\dot x+\dot x_{\mathrm{a}}^\top
      \bigl[\tau-g_{\mathrm{a}}(x^d)+K_{\mathrm{p}}(x_{\mathrm{a}}-x_{\mathrm{a}}^d)\bigr]\\
    &=-\dot x^\top D_x\dot x-\dot x_{\mathrm{a}}^\top K_{\mathrm{d}}\dot x_{\mathrm{a}}\le0.
\end{aligned}
\end{equation}
Since $D_x\succ0$, $\dot V=0$ requires $\dot x=0$; remaining there requires $\nabla W(x)=0$, hence $x=x^d$ by strong convexity.
The compact positively invariant sublevel sets of $V$ establish Lyapunov stability and allow LaSalle's invariance principle to give global convergence to $(x^d,0)$.